\documentclass{article}
\newif\ifarxiv
\arxivtrue
\usepackage{iclr2027_conference,times}
\ifarxiv\iclrfinalcopy\fi
\ifarxiv
  \newcommand{\coderepo}{https://github.com/CSSLab/Tandem-RLVR}
\else
  \newcommand{\coderepo}{https://anonymous.4open.science/r/Tandem-RLVR-86BB/}
\fi
\usepackage[utf8]{inputenc}
\usepackage[T1]{fontenc}
\usepackage{hyperref}
\usepackage{url}
\usepackage{booktabs}
\usepackage{amsfonts}
\usepackage{nicefrac}
\usepackage{microtype}
\usepackage{xcolor}
\usepackage{amsmath}
\usepackage{graphicx}
\usepackage{algorithmic}
\usepackage{algorithm}
\usepackage{mathtools}
\usepackage{amsthm}
\usepackage{dsfont}
\usepackage{multirow}
\usepackage[all]{nowidow}
\newcommand{\xhdr}[1]{\vspace{2mm}\noindent{{\bf #1.}}}
\newcommand{\modelname}{TRLVR}
\hypersetup{pdftitle={Tandem Reinforcement Learning with Verifiable Rewards},pdfauthor={Difan Jiao, Raghav Singhal, Robert West, Ashton Anderson}}
\title{Tandem Reinforcement Learning with \\ Verifiable Rewards}
\author{%
  Difan Jiao\,$^{\dagger *}$ \quad
  Raghav Singhal\,$^{\ddagger}$ \quad
  Robert West\,$^{\ddagger}$ \quad
  Ashton Anderson\,$^{\dagger *}$ \\[6pt]
  {\small $^{\dagger}$\,University of Toronto \qquad $^{\ddagger}$\,EPFL} \\[2pt]
  {\small $^{*}$\,Contact: \texttt{\{difanjiao, ashton\}@cs.toronto.edu}}
}
\begin{document}
\maketitle
\ifarxiv\lhead{Preprint. Under review.}\else\lhead{Under review as a conference paper at ICLR 2027}\fi
\begin{abstract}
Reinforcement learning with verifiable rewards (RLVR) has driven substantial gains in language-model reasoning. However, models fine-tuned with RLVR need not remain compatible with weaker partners: gains achieved alone may erode under collaboration, while policy drift may place reasoning beyond a weaker model's predictive reach. To address both problems, we introduce Tandem RLVR (\modelname{}), bringing the recently proposed tandem-training paradigm beyond proof-of-concept settings and into modern RLVR. In \modelname{}, a \emph{senior} learns from rewarded trajectories co-generated with a frozen \emph{junior}, while the standard GRPO objective is applied only to senior-emitted tokens. Training Qwen3-4B-Instruct on competition math, \modelname{} performs on par with a matched GRPO senior when reasoning alone. We characterize compatibility through the complementary lenses of handoff robustness and junior legibility. Under reasoning-step handoffs, the \modelname{} senior retains nearly all of its solo performance with the junior, reducing the \emph{communication tax} to a negligible loss. The senior's reasoning is also more legible to the junior: its token distribution stays more closely anchored to the junior's, and the junior predicts the senior's chain-of-thought more readily token by token. An ablation that explicitly regularizes GRPO toward the junior does not recover the same combination of capability and compatibility. These results provide initial evidence that tandem rollout structure is a promising direction for keeping RLVR reasoning gains accessible to weaker partners.\footnote{The codebase is available at \url{\coderepo}.}
\end{abstract}

\section{Introduction}

Reinforcement learning with verifiable rewards (RLVR) has emerged as a dominant post-training paradigm in language-model post-training. In RLVR, a model samples a solution, an external verifier scores the final answer, and policy optimization reinforces successful trajectories. It effectively produces large gains in reasoning performance across various challenging domains~\citep{shao2024deepseekmath, guo2025deepseek, yu2025dapo}.


However, this objective rewards correct answers without requiring reasoning that weaker agents can follow. As a model becomes better at solving problems, its reasoning can become harder for its pre-RL base model, weaker partners, or human overseers to predict, continue, or understand. Models that excel alone can fail to retain that performance when collaborating with weaker partners~\citep{davidson2025collaboration}. RLVR also shifts the policy away from its base distribution and, in some reported cases, produces idiosyncratic token patterns rather than transparent explanations~\citep{guo2025deepseek,kirchner2024prover,karpov2025steganographic,skaf2025large}.

This incompatibility need not reflect a defect in capability, since token-level analyses attribute part of RLVR's reasoning gains to precisely the positions where the trained policy departs furthest from the base~\citep{li2025impact,meng2026sparse}. This suggests that the model does not reason well in spite of its private dialect but partly because of it; the illegibility is tied to the capability itself~\citep{jose2025illegible}. GPT-6 Astra, for example, solves ARC-AGI-3 games while planning in an algebraic shorthand of its own invention~\citep{arcprize2026astra}. Reasoning traces extracted from deployed frontier models also reveal stretches of obfuscated reasoning in challenging tasks~\mbox{\citep{panfilov2026stealing}}. OpenAI describes this widening gap between capability and human understanding as an ``alien mind,'' arguing that model behavior becomes harder to interpret as capability grows~\citep{pachocki2026alien}.

There are thus two reasons for concern. We build AI for collaboration with people and weaker systems~\citep{nasem2022humanai,tang2026maia4all}, but a driving policy, for example, can outperform the driver it assists and still be unsafe when that driver cannot anticipate it~\citep{teoh2017rage}. We also want to learn from AI beyond merely delegating work to it~\citep{shin2023superhuman,tang2024maia}. Knowledge transfer and oversight depend on being able to make sense of a model's reasoning. A chain of thought that no one can parse offers nothing to learn from, and it can also make weak-to-strong supervision~\citep{burns2023weak} more difficult.

\begin{figure}[t]
  \centering
  \includegraphics[width=\linewidth]{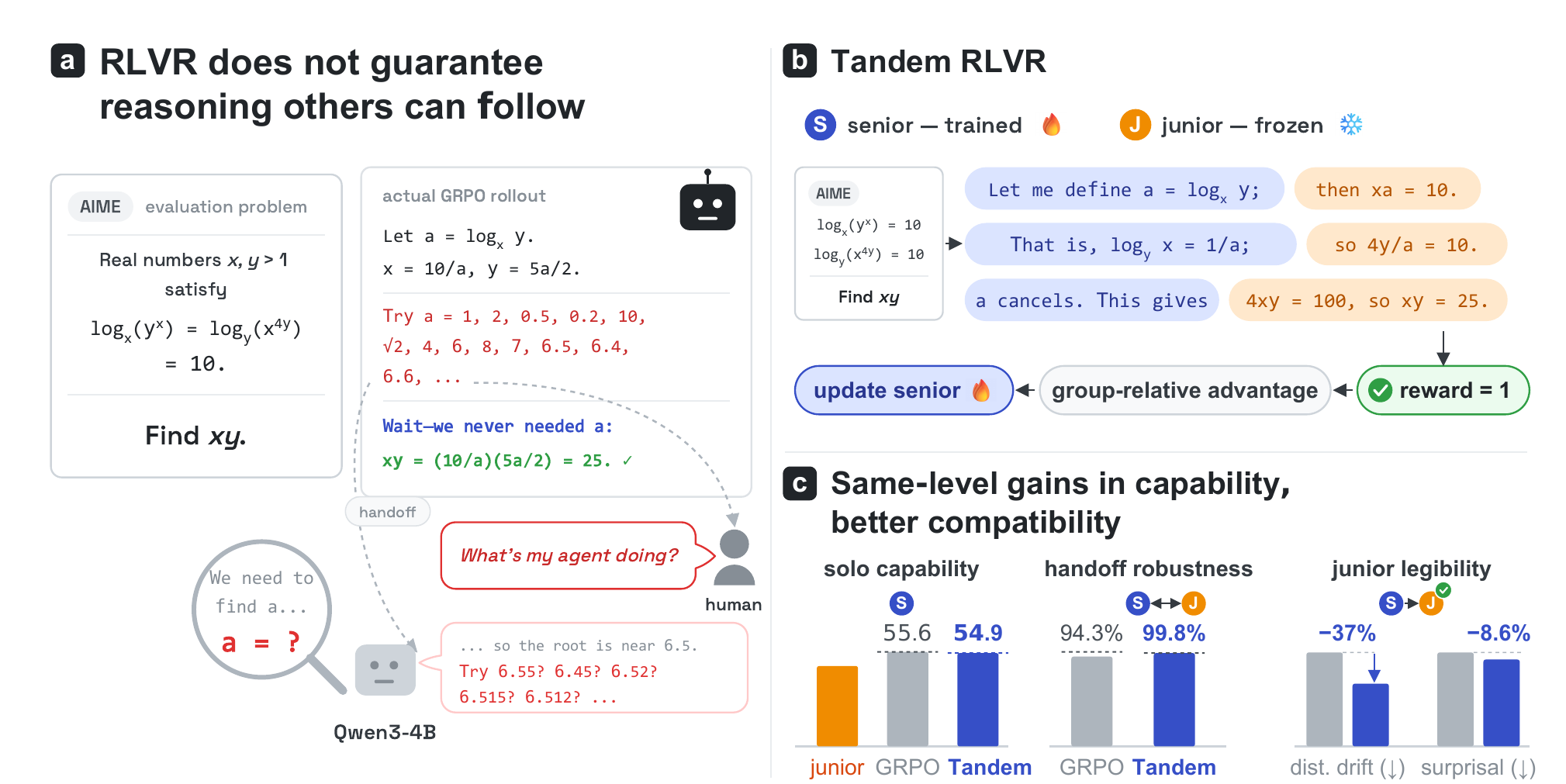}
  \vspace{-0.5cm}
  \caption{ (a) In an AIME example from our evaluation, a GRPO-trained model repeatedly guesses an unnecessary intermediate value before realizing that it cancels, producing a chain of thought that weaker partners and overseers find hard to follow. (b) A glance at our methodology, Tandem RLVR (TRLVR). (c) On competition math, the Tandem RLVR senior matches GRPO in solo reasoning capability, retains nearly all of that capability when the junior takes over half the reasoning, and expresses its reasoning in language the junior can follow.}
  \label{fig:trl}
\end{figure}

We ask whether RLVR can improve capability without eroding compatibility. Tandem training, a recently introduced paradigm, offers a way to train for compatibility: a trainable \emph{senior} model co-generates each rollout with a frozen \emph{junior} partner and learns from their joint reward~\citep{west2026tandem}. Proofs of concept exist in chess~\citep{hamade2024designing} and a simplified LLM reasoning setting~\citep{west2026tandem}, but it remains unclear whether tandem training can preserve the capability gains of modern RLVR while making those gains accessible to weaker partners. In this work, we adapt tandem training to RLVR as Tandem RLVR (\modelname{}). The senior and junior start from the same base model and co-generate each response by stochastically alternating at word boundaries (Figure~\ref{fig:trl}b). A binary verifier then scores the completed response, and the standard GRPO objective updates the senior on its own tokens. The reward, verifier, and policy-gradient loss are unchanged from the GRPO baseline, isolating rollout structure as the training difference.



With Qwen3-4B-Instruct on competition math, \modelname{} preserves the capability gains of RLVR: its senior performs on par with a matched GRPO senior when reasoning alone. We then assess compatibility along two complementary dimensions: \textit{handoff robustness} and \textit{junior legibility}. Handoff robustness measures how much of the senior's solo performance is retained when a weaker partner joins the reasoning process. Under reasoning-step handoffs, the \modelname{} senior retains $99.8\%$ of its solo pass@$4$ on macro average, compared with $94.3\%$ for its GRPO counterpart, reducing the \emph{communication tax} to a negligible loss~\citep{davidson2025collaboration}. Junior legibility measures how well the junior can predict the senior's reasoning. Compared with GRPO, the \modelname{} senior stays closer to the junior's token distribution, and its reasoning traces are more predictable to the junior token by token. An ablation that explicitly regularizes GRPO toward the junior does not recover the same combination of capability and compatibility.

Taken together, these findings suggest that tandem rollouts are a promising direction for keeping RLVR gains accessible to weaker partners. We view \modelname{} as an initial step toward training for both handoff robustness and junior legibility alongside capability, with a broad design space left to explore.

\section{Related work}
\label{sec:related}

\xhdr{Tandem training} \citet{hamade2024designing} introduced tandem training in chess, showing that training a model to win alongside a weaker partner differs from training it to win alone. \citet{west2026tandem} extended tandem training to language modeling. On GSM8K, randomized handoffs to a frozen junior during RL taught the senior to reduce jargon and adapt its language while preserving task accuracy. These studies did not evaluate tandem training with the long chains of thought and policy optimization involved in modern RLVR. We study tandem training in this modern RLVR setting with \modelname{} and evaluate both solo capability and compatibility, using handoff robustness and junior legibility to measure the compatibility.

\xhdr{Compatibility concerns in RLVR} RLVR rewards verified task success without requiring either that the learned capability remain usable when a weaker partner joins the reasoning process or that the reasoning trace remain within that partner's predictive reach. More broadly, models that perform well alone can degrade sharply in heterogeneous collaboration, a phenomenon termed the \emph{collaboration gap}~\citep{davidson2025collaboration,wynn2025talk}. Within RLVR, DeepSeek-R1-Zero exhibits reduced readability and unprompted language mixing~\citep{guo2025deepseek}, and RLVR can induce Chinese--English code-switching in bilingual reasoners~\citep{li2025impact}; related work further identifies steganographic reasoning patterns inaccessible to downstream monitors~\citep{karpov2025steganographic,skaf2025large,baker2025monitoring,drori2026output}. Recent evaluations of weak-to-strong legibility likewise find that strong reasoning performance need not produce traces useful to weaker models~\citep{roytburg2026measuring,jose2025illegible,li2025small}. Yet not all distribution shift is pathological: token-level interventions indicate that a small set of high-divergence positions can carry RLVR's capability gains~\citep{meng2026sparse}. The target is therefore not to suppress drift indiscriminately, but to preserve capability while keeping it accessible to weaker partners.

\xhdr{Approaches to compatibility and legibility} When the desired behavior can be specified in advance, textual principles can prescribe it~\citep{bai2022constitutional}, supervised reasoning traces can instantiate a target solution style~\citep{magister2023teaching,li2025small,cetin2025teachers}, and process supervision can reward valid intermediate steps~\citep{uesato2022solving,lightman2023let,anwar2026monitorability,surana2026rollout}. Prover--verifier games train for solution checkability by a learned verifier~\citep{kirchner2024prover}. These approaches are complementary to tandem training. Specification-based methods are well suited when a target behavior is available, while prover--verifier methods operationalize legibility as checkability by a designated verifier. Tandem training uses a different interaction-based criterion: whether a designated weaker generator can take over the senior's partial trajectory and still complete the task successfully. Because the task verifier rewards the co-generated trajectory as a whole, tandem training requires neither a canonical surface form nor additional process labels.

\section{Tandem RLVR}

\subsection{Preliminaries: tandem training}
\label{sec:prelim}

Tandem training~\citep{hamade2024designing,west2026tandem} is a recently introduced paradigm for training models that remain effective when paired with a partner. It uses task performance on jointly generated rollouts as the training signal, so compatible reasoning need not be specified in advance. By comparison, explicit methods such as system prompts or supervised finetuning on canonical solutions require an \emph{a priori} specification that depends on deployment context~\citep{west2026tandem,bai2022constitutional,lightman2023let}.

A trainable \emph{senior} policy $\pi_{\text{sen}}$ and a frozen \emph{junior} policy $\pi_{\text{jun}}$ stochastically alternate generations to produce each rollout. At predetermined handoff boundaries such as tokens, words, or sentences, a coin flip determines the next active model. For example, token-level handoffs use a new coin flip for each token. A single reward function scores the completed rollout as one trajectory, and a policy-gradient algorithm updates the senior using that reward while the junior remains unchanged. However, success under these handoffs does not by itself establish that the junior semantically understands the senior's reasoning. Task success may arise from information uptake, division of labor, error correction, or partially independent reasoning.

In \modelname{}, we initialize the senior and junior from the same base model (\S\ref{sec:selfpair}) and co-generate rollouts with word-level handoffs (\S\ref{sec:rollout}). The senior is then updated with Group Relative Policy Optimization (GRPO)~\citep{shao2024deepseekmath}, with the loss applied only to senior-emitted tokens (\S\ref{sec:loss}).

\subsection{Tandem pairs}
\label{sec:selfpair}

We use the senior's own pre-RL initialization as the junior to study whether RLVR can improve capability while preserving compatibility with the original policy. The two models begin as identical copies, but the senior becomes stronger through RLVR while the junior remains frozen. The capability gap therefore develops during training. This lets us examine how RLVR changes compatibility without introducing an initial difference in capabilities or reasoning style between the models. Because each reward is computed on a jointly generated response, the senior's updates continue to depend on how well it works with the frozen junior as its capabilities improve. Our goal is for the senior to improve its capabilities while minimizing drift from its initial state, so that it remains compatible and intelligible to the junior. We therefore initialize $\pi_{\text{sen}}$ and $\pi_{\text{jun}}$ from the same base model and freeze $\pi_{\text{jun}}$ for the duration of training.

\subsection{Tandem rollout}
\label{sec:rollout}

A tandem rollout produces a shared response $y_{1:T}$ from prompt $x$, where each step $t$ is generated by an active model $a_t \in \{\text{sen}, \text{jun}\}$. At each step, both models receive the context $x \cdot y_{<t}$, the active model samples $y_t \sim \pi_{a_t}(\cdot \mid x \cdot y_{<t})$, and the chosen $y_t$ is appended and then fed to both models so that both states condition on the same history. Following~\citet{west2026tandem}, the active model is redrawn by an independent fair coin flip at every word boundary: if $y_t$ begins a new orthographic word, $a_{t+1}$ is set to the senior with probability $p$ and to the junior otherwise; if $y_t$ does not begin a new word, $a_{t+1} = a_t$. We use $p = 0.5$ in consistency with original tandem training. Details of this setup are included in Appendix~\ref{app:rollout}, alongside a summary of the method in Figure~\ref{fig:glance}.

We choose to stochastically alternate at word-level granularity over alternative choices (e.g., token-level or sentence-level). First, at sentence- or paragraph-level the senior can dominate a rollout simply by suppressing boundary tokens (e.g., end-of-sentence or end-of-line tokens) that would force a potential handoff\footnote{This reward-hacking failure mode is induced by the schedule itself: as the senior strengthens, suppressing boundary tokens becomes a low-cost route to extending its own control and avoiding handoffs.}, collapsing the schedule back to GRPO. Second, at token level, alternation at subword granularity disrupts coherent utterances before they form. In contrast, word-level boundaries fire on every natural-language word and are difficult to suppress without sacrificing fluency, making them the coarsest granularity the senior cannot trivially erase by output choice.

\subsection{Tandem policy update}
\label{sec:loss}

Each tandem rollout receives a single binary reward $r(y) \in \{0, 1\}$ from a math verifier, computed on the full response $y_{1:T}$ regardless of which model emitted each token. The senior is updated with the standard GRPO objective~\citep{shao2024deepseekmath}, with the response mask set to select only senior-emitted positions; the per-token policy-gradient contribution is

\begin{equation}
  \mathcal{L}_{\text{\modelname{}}}(sen) \;=\; -\,\mathbb{E}\!\left[\;\sum_{t \,:\, a_t = \text{sen}} A_t \,\log \pi_{\text{sen}}(y_t \mid x \cdot y_{<t})\;\right],
\end{equation}

where $A_t$ is the standard GRPO group-relative advantage; the clipped surrogate ratio and KL term are also identical to GRPO and omitted here for brevity. Junior-emitted tokens enter the response, and so affect the reward the senior receives, but contribute no gradient.

Specifically, while the original tandem-training formulation also includes a soft junior-imitation term that pulls $\pi_{\text{sen}}$ toward $\pi_{\text{jun}}$ on junior-emitted positions with coefficient $\lambda_{\text{jun}}$, we set $\lambda_{\text{jun}} = 0$. Beyond removing a hyperparameter we have no principled value for, this keeps the per-token loss formally the same as GRPO. Thus, the only difference between the \modelname{}  and GRPO objectives is the rollout structure that produces $y$, not an auxiliary regularization term. We evaluate each run at its best validation checkpoint under its own rollout protocol, and also compare the two at a matched training step.

The end-to-end \modelname{} pipeline is illustrated in Algorithm~\ref{alg:trl} in Appendix~\ref{app:algorithm}.

\section{Results}
\label{sec:results}

\begin{figure}[t]
  \centering
  \includegraphics[width=\linewidth]{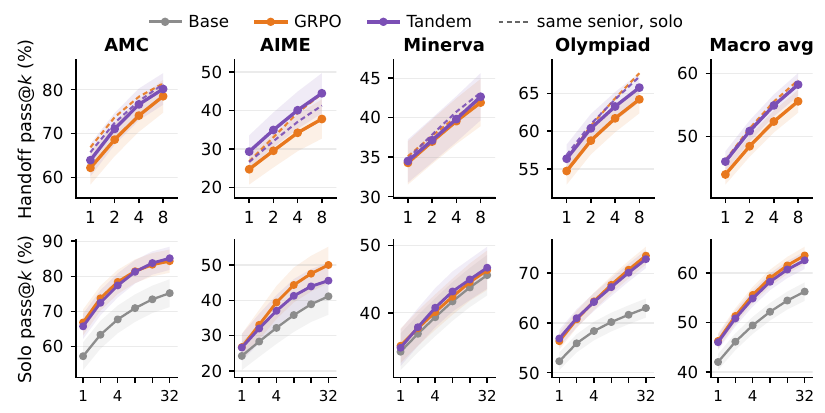}
  \caption{Reasoning-step handoff pass@$k$ (top) and solo pass@$k$ (bottom) of Qwen3-4B-Instruct and its GRPO- and \modelname{}-trained seniors on competition math benchmarks ($\uparrow$).}
  \label{fig:solo_handoff}
\end{figure}

\subsection{Experimental setup}
\label{sec:setup}

We fine-tune Qwen3-4B-Instruct-2507~\citep{yang2025qwen3} on DeepScaleR~\citep{tan2025deepscaler} with a binary correctness reward on the boxed final answer. Our baseline is GRPO~\citep{shao2024deepseekmath} on the same base model, with optimization settings shared with \modelname{} (Appendix~\ref{app:hparams}). We evaluate on AMC~23--25, AIME~24--26, Minerva Math~\citep{lewkowycz2022solving}, and OlympiadBench~\mbox{\citep{he2024olympiadbench}}, and report pass@$k$ via an unbiased estimator following \citet{chen2021evaluating}. We estimate bootstrap standard errors by resampling evaluation problems, reported as $\pm$ values in tables and as shaded bands in figures.

This section is organized in three parts. In \S\ref{sec:eval-rawcap}, we measure the resulting models' solo reasoning capabilities, examining whether the tandem rollout preserves the reasoning gains delivered by GRPO. We then turn to handoff robustness in \S\ref{sec:eval-handoff}, the central behavioural property tandem training was originally proposed to elicit; we pair the trained models with the junior model at inference time and evaluate the team's performance. Finally, in \S\ref{sec:eval-drift} we ask whether the senior's chain-of-thought stays legible to the junior, measuring both distributional drift from the base and how predictable the junior finds the chain-of-thought token by token.


\subsection{\modelname{} retains the reasoning gains of GRPO}
\label{sec:eval-rawcap}

We first measure each model's solo reasoning capability, with no junior in the loop at test time, to test whether the tandem rollout structure preserves the gains of standard RLVR training. Each senior generates $n$ independent samples per problem; we report pass@$k$ via the unbiased estimator for $k \in \{1, 2, 4, 8, 16, 32\}$.

The bottom row of Figure~\ref{fig:solo_handoff} shows pass@$k$ curves for the pre-RL base and the seniors trained with GRPO and \modelname{} across benchmarks and their macro average. Two observations stand out. First, both trained seniors outperform the base model across the full $k$ range on all four benchmarks, confirming that RLVR delivers a non-trivial capability gain. Second, the GRPO and \modelname{} curves track each other closely throughout, with overlapping bootstrap uncertainty bands and no consistent direction across benchmarks. Together, these results provide empirical evidence that \modelname{} preserves the \emph{solo} reasoning gains achieved by GRPO.

\subsection{\modelname{} offers better handoff robustness}
\label{sec:eval-handoff}

\begin{figure}[t]
  \centering
  \includegraphics[width=\linewidth]{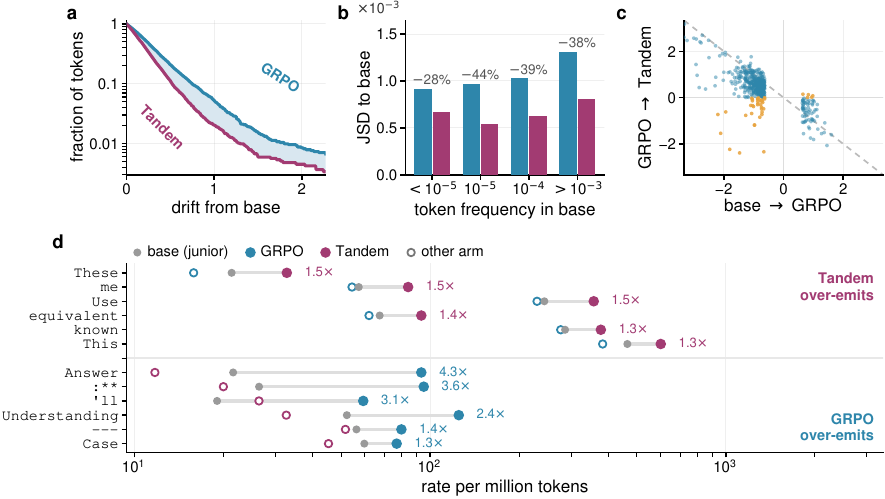}
  \caption{Distributional deviation of the GRPO- and \modelname{}-trained seniors from the junior: (a) survival curve of the absolute per-token log-ratio, (b) Jensen--Shannon divergence to the junior by token frequency in the junior, (c) the GRPO senior's token-wise displacement from the junior against the \modelname{} senior's displacement from GRPO, for the tokens the GRPO senior displaces most, and (d) emission rate of the tokens each senior over-emits most.}
  \label{fig:drift_and_distance}
\end{figure}

In \S\ref{sec:rollout} we argued against training with sentence- or paragraph-level handoffs, on the grounds that the senior could game such schedules by suppressing the boundary tokens that would force a switch. For evaluation, however, there is no risk of this type of gaming behavior, and we want to measure how cooperation between language models actually unfolds in practice. In multi-agent reasoning systems, such as drafter-target setups for inference and human-AI handoffs, the natural unit of communication is a reasoning step rather than an individual word, with one party producing a step and a partner picking up where it left off~\citep{chu2025ssr, wang2025mars, davidson2025collaboration}\footnote{also resembling a \emph{ply} of chess games in~\citet{hamade2024designing}.}. This motivates our evaluation of handoff robustness under a reasoning-step schedule: at inference, the senior and the junior alternate at every \texttt{\textbackslash n\textbackslash n}-boundary token, each producing one reasoning step before yielding control to the other, with the team-produced response scored as a whole.

The top row of Figure~\ref{fig:solo_handoff} shows handoff pass@$k$ against each senior's own solo pass@$k$, drawn dashed, so the gap between a senior's two curves is what handing off costs it. \modelname{} leads on all four benchmarks and at every $k$. The margin is widest on AIME, where it reaches $+6.6$ points at $k{=}8$, and on macro average it widens with $k$ throughout.
We call the team's pass@$k$ \emph{handoff accuracy}, and measure alongside it \emph{handoff robustness}: handoff accuracy as a percentage of the same senior's solo pass@$k$, computed from the macro-averaged accuracies. The ratio measures retention of solo performance under handoffs, though it is read alongside solo capability, since a senior that scores lower alone has less to lose. At $k{=}4$ the GRPO senior retains $94.3\%$ of its solo score and the \modelname{} senior $99.8\%$. Together with the solo results in \S\ref{sec:eval-rawcap}, these findings show that the \modelname{} senior is no less capable alone than the GRPO senior and barely loses any of that capability when the junior takes over half of the reasoning. We report paired $t$-tests on AMC, AIME, and Minerva for this gap, and for the further \modelname{} gains reported in the following subsection, in Appendix~\ref{app:significance}.

\subsection{\modelname{} keeps the senior's chain-of-thought legible to the junior}
\label{sec:eval-drift}

\begin{table}[t]
\centering
\small
\setlength{\tabcolsep}{4pt}
\caption{Junior legibility of senior generated chain-of-thoughts
(Qwen3-4B-Instruct junior). \textbf{Bold} marks the better model per cell.}
\label{tab:legibility}
\begin{tabular}{@{}llccccc@{}}
\toprule
Metric & Model & AMC & AIME & Minerva & Olympiad & Avg. \\
\midrule
\multirow{3}{*}{CE (nats, $\downarrow$)}
  & GRPO & $0.119_{\pm 0.004}$ & $\mathbf{0.147}_{\pm 0.006}$ & $0.113_{\pm 0.003}$ & $0.139_{\pm 0.003}$ & $0.129_{\pm 0.002}$ \\
  & \modelname{} & $\mathbf{0.107}_{\pm 0.005}$ & $\mathbf{0.147}_{\pm 0.007}$ & $\mathbf{0.093}_{\pm 0.003}$ & $\mathbf{0.126}_{\pm 0.003}$ & $\mathbf{0.118}_{\pm 0.002}$ \\
  & $\Delta$ & $\mathbf{-0.012}_{\pm 0.006}$ & $+0.000_{\pm 0.009}$ & $\mathbf{-0.020}_{\pm 0.004}$ & $\mathbf{-0.013}_{\pm 0.004}$ & $\mathbf{-0.011}_{\pm 0.003}$ \\
\midrule
\multirow{3}{*}{$\alpha$ (\%, $\uparrow$)}
  & GRPO & $95.89_{\pm 0.12}$ & $95.38_{\pm 0.15}$ & $96.08_{\pm 0.10}$ & $95.40_{\pm 0.07}$ & $95.69_{\pm 0.06}$ \\
  & \modelname{} & $\mathbf{97.29}_{\pm 0.07}$ & $\mathbf{96.87}_{\pm 0.09}$ & $\mathbf{97.53}_{\pm 0.05}$ & $\mathbf{97.06}_{\pm 0.04}$ & $\mathbf{97.19}_{\pm 0.03}$ \\
  & $\Delta$ & $\mathbf{+1.40}_{\pm 0.14}$ & $\mathbf{+1.48}_{\pm 0.17}$ & $\mathbf{+1.45}_{\pm 0.11}$ & $\mathbf{+1.66}_{\pm 0.08}$ & $\mathbf{+1.50}_{\pm 0.06}$ \\
\bottomrule
\end{tabular}
\end{table}

We now ask whether the senior's chain-of-thought stays legible to the junior, in two senses. The first is distributional, whether the senior's overall token usage stays near the junior's or drifts away as documented for RLVR at scale (\S\ref{sec:related}). The second is sequential, whether the junior, reading the senior's chain-of-thought left to right, finds each next token predictable. The two measures capture different aspects of legibility, since similar marginal token distributions do not necessarily make the senior's next token predictable to the junior. During tandem rollouts, the frozen junior can take over at word boundaries, so its continuations affect the reward used to update the senior. We therefore expect training to favor reasoning that stays closer to the junior's distribution and is more predictable to it.

\subsubsection{\modelname{} curbs distributional drift}

We estimate each senior's marginal token distribution from its generations on the evaluation sets, and measure per-token deviation from the junior by the log-ratio $\log(p_\mathrm{sen}(t)\,/\,p_\mathrm{jun}(t))$; Jensen--Shannon divergence (JSD) to the junior aggregates these deviations, weighted by how often each token is actually used. Panels (a) and (b) of Figure~\ref{fig:drift_and_distance} show that the \modelname{} senior stays closer to the junior's token distribution than the GRPO senior. Its survival curve lies below GRPO's at every deviation threshold, its JSD to the junior is 37\% lower, and the gap persists in every band when the divergence is decomposed by token frequency in the junior, so the reduction comes neither from a few common tokens nor from the long tail. Panel (c) makes the comparison token by token: among the top tokens GRPO displaces most, 91\% move back toward the junior under \modelname{}, and the two moves are anti-correlated (Spearman $\rho = {-}0.60$), so the further GRPO displaces a token, the more of that displacement \modelname{} tends to undo.

Panel (d) provides a case study of the most displaced tokens under each recipe. From the tokens each senior over-emits we keep the interpretable ones, ranked by how strongly they are over-emitted. GRPO's extra mass goes to the markup of a formatted report, headings, bold labels, and horizontal rules; the \modelname{} senior's extra mass goes to connective phrases that carry one step into the next. Both methods reward the final answer, but in \modelname{} the junior receives the senior's intermediate reasoning as context for its own continuation. This may help explain why \modelname{} more often produces phrases that connect successive reasoning steps. Appendix~\ref{app:token-drift} extends the case study to the phrase level.

\subsubsection{\modelname{} reduces the junior's surprisal}
\label{sec:eval-legibility}

We measure the sequential sense with two analyses, summarized in Table~\ref{tab:legibility}. The first is the junior's per-token cross-entropy on the senior's chain-of-thought: the average \emph{nats} required to encode each senior token under the junior's predictive distribution, where lower means the junior is less surprised. The \modelname{} senior's cross-entropy is lower by $0.011$ nats on average, with clear gains on AMC, Minerva, and OlympiadBench and a match on AIME. This corresponds to an $8.6\%$ reduction in per-token surprisal on average, reaching $17\%$ on Minerva.

Appendix~\ref{app:ce-controls} provides two controls on this metric. Lower junior cross-entropy is associated with correctness within the same problem, and the \modelname{} advantage grows after subtracting each trace's cross-entropy under its generating senior, ruling out greater predictability under the senior as the explanation for the gap.

The second is the token-level distribution overlap $\alpha = \sum_v
\min(p_\mathrm{sen}(v\mid\mathrm{\cdot}),\,p_\mathrm{jun}(v\mid\mathrm{\cdot}))$ averaged over all positions \citep{leviathan2023fast}: the probability mass both models assign to the same tokens at each step, ranging from $0$ (fully disjoint) to $1$ (identical), and reported as a percentage. As shown, the \modelname{} senior's distribution overlap is uniformly higher across all four benchmarks. On average, the overlap increases by $1.50$ percentage points, from $95.69\%$ for GRPO to $97.19\%$ for \modelname{}. The remaining gap to full overlap falls from $4.31\%$ to $2.81\%$, a reduction of roughly $35\%$.

\vspace{1.2mm}
\noindent\colorbox{gray!12}{\parbox{\dimexpr\linewidth-2\fboxsep\relax}{
\textbf{Section takeaway.} The \modelname{} senior retains the solo reasoning gains of GRPO (\S\ref{sec:eval-rawcap}), preserves nearly all of that performance when the junior takes over half the reasoning (\S\ref{sec:eval-handoff}), and keeps its reasoning in language the junior can follow (\S\ref{sec:eval-drift}).}}

\section{Discussion}

\subsection{Training dynamics of \modelname{}}

\begin{figure}[t]
    \centering
    \includegraphics[width=\linewidth]{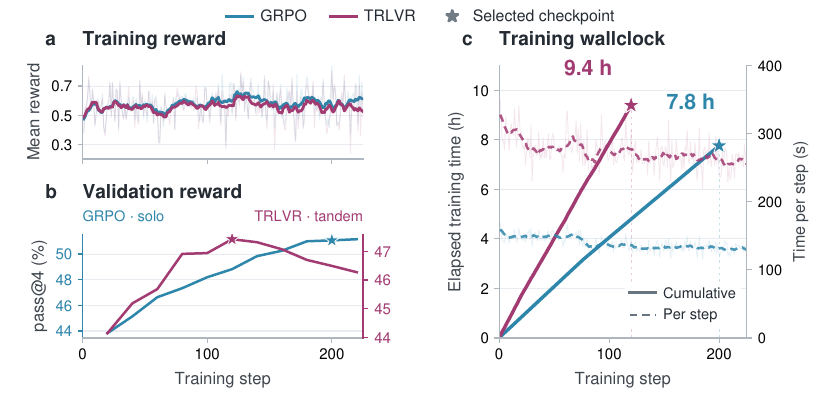}
    \vspace{-0.5cm}
    \caption{Training dynamics of \modelname{} and GRPO.
    (a)~Mean reward.
    (b)~Validation pass@$4$ under each run's evaluation protocol.
    (c)~Accumulated wallclock to selected checkpoints (solid, left axis) and per-step wallclock (dashed, right axis).}
    \label{fig:training_dynamics}
\end{figure}

\modelname{} generates each rollout with two models, which raises questions about training stability and computational cost. Figure~\ref{fig:training_dynamics} reports training reward, validation performance, and wallclock time. For these runs, we select \modelname{} checkpoints by tandem-rollout validation pass@$4$ and GRPO checkpoints by solo validation pass@$4$. This keeps \modelname{} checkpoint selection consistent with its cooperative training objective. Panel~(b) therefore uses separate axes for the two validation curves, which measure performance under different rollout protocols.

Panel~(a) shows that training reward follows a stable trajectory under \modelname{}, closely tracking GRPO throughout.
These reward curves suggest that tandem rollouts preserve stable learning in this setting. Running both models at each generation step does add computational cost. Our initial implementation could not support the long contexts required for RLVR, so we extended the vLLM backend~\citep{kwon2025vllm} to run tandem generation within its inference engine (Appendix~\ref{app:rollout}). Each model maintains its own paged KV cache for the shared response, avoiding prefix recomputation at handoffs. The selected \modelname{} checkpoint is at step $120$, compared with step $200$ for GRPO, partially offsetting the per-step overhead. Training time to these checkpoints is 9.4 hours for \modelname{} versus 7.8 hours for GRPO (panel~c).

We also evaluate GRPO at step 120, the step of the selected \modelname{} checkpoint (matched step in Table~\ref{tab:ablation}). At the matched step, \modelname{} scores higher alone and under handoff, with handoff robustness of $99.8\%$ versus $95.2\%$. Junior cross-entropy is slightly lower for GRPO there, as is capability: within the GRPO run, shorter training trades capability for legibility. \modelname{} improves on that trade-off, being more capable than GRPO at the matched step, more legible than GRPO at its best validation step, and more robust under handoff than either.




\subsection{Ablation study: KL-regularization towards the junior}
\label{sec:ablation}

We next test whether policy-level KL regularization toward the junior achieves the anchoring that \modelname{} obtains through its rollout structure. \modelname{} anchors the senior only implicitly, through co-generation under a shared reward, so an explicit per-token KL penalty toward the same junior is the natural ablation. In our setup this baseline is exactly standard KL-regularized GRPO: the recipe regularizes toward the pre-RL reference policy, which is the same frozen checkpoint that serves as \modelname{}'s junior. We denote this baseline \emph{KL-Reg} and train it at the documented default coefficient of $\beta=10^{-3}$ and at $\beta=10^{-1}$, using the same training data, optimizer settings, and evaluation protocol as \modelname{} and GRPO.

Table~\ref{tab:ablation} reports macro averages for solo capability, handoff accuracy, and junior legibility, together with handoff robustness. At the default strength, KL-Reg preserves solo capability but leaves handoff accuracy and robustness at the GRPO level, without improving junior legibility. At $\beta=10^{-1}$, KL-Reg has the lowest junior cross-entropy in the table, but solo capability drops by more than five percentage points relative to GRPO and handoff accuracy is also the lowest. Its $98.0\%$ handoff robustness indicates that it retains most of this reduced solo capability. In comparison, \modelname{} retains $99.8\%$ under handoff while preserving GRPO-level solo capability.

These results are consistent with the difference in how the two objectives use the junior. KL-Reg measures next-token divergence from the junior on the senior's own rollouts and receives task rewards from senior-only responses. In \modelname{}, junior continuations form part of the rewarded trajectory, so the senior's updates depend on how the pair completes the task. Appendix~\ref{app:ablation} details this difference between the training signals.

\begin{table}[t]
\centering
\setlength{\tabcolsep}{6pt}
\caption{GRPO at its best validation step and at \modelname{}'s matched step, KL-regularized GRPO at two coefficients, and \modelname{} on the evaluation axes of \S\ref{sec:results} (macro averaged).}
\label{tab:ablation}
\small
\begin{tabular}{l cccc}
\toprule
\multirow{2}{*}{Method} & Solo Capability & Handoff Accuracy & Handoff Robustness & Junior Legibility \\
& {\small (pass@$4$, \%, $\uparrow$)} & {\small (pass@$4$, \%, $\uparrow$)} & {\small (\%, $\uparrow$)} & {\small (nats, $\downarrow$)} \\
\midrule
GRPO & $55.56_{\pm 2.1}$ & $\underline{52.40}_{\pm 2.3}$ & $94.3$ & $0.129_{\pm 0.002}$ \\
GRPO (matched step) & $53.14_{\pm 1.7}$ & $50.60_{\pm 1.8}$ & $95.2$ & $\underline{0.114}_{\pm 0.002}$ \\
KL-Reg, $\beta{=}10^{-3}$ & $55.24_{\pm 2.2}$ & $52.09_{\pm 2.3}$ & $94.3$ & $0.131_{\pm 0.003}$ \\
KL-Reg, $\beta{=}10^{-1}$ & $50.15_{\pm 2.3}$ & $49.15_{\pm 2.3}$ & $\underline{98.0}$ & $\mathbf{0.114}_{\pm 0.003}$ \\
\modelname{} & $54.85_{\pm 2.1}$ & $\mathbf{54.74}_{\pm 2.4}$ & $\mathbf{99.8}$ & $0.118_{\pm 0.002}$ \\
\bottomrule
\end{tabular}
\end{table}

\section{Conclusion}

RLVR rewards solving problems alone and leaves open whether the gains survive a weaker partner joining the reasoning. We introduced \modelname{}, which co-generates each rewarded trajectory with a frozen copy of the pre-RL base and applies the standard GRPO objective to senior tokens only, leaving reward, verifier, and loss unchanged. Across four competition-math benchmarks, the senior matches GRPO alone, keeps $99.8\%$ of its macro-average solo pass@$4$ under reasoning-step handoffs against $94.3\%$ for GRPO, and stays more legible to the junior, in token distribution and in next-token predictability.


Several directions for future work are worth highlighting. First, replacing the fixed self-paired junior with a pool of diverse juniors varying in capability, style, language, or tool use could regularize the senior toward more broadly intelligible behavior. Second, pairing the senior with juniors calibrated to specific human skill levels would provide a more direct bridge to the genuine human-AI handoff scenarios that originally motivated the tandem framework. We view this direction as currently bottlenecked by the absence of comparably well-calibrated, human-skill-level proxies for general-purpose language models, and a promising avenue once such models mature.

\section*{Acknowledgments}

This research is funded by grants from the Natural Sciences and Engineering Research Council of Canada (NSERC), the Canada Foundation for Innovation, and the Ontario Research Fund. We are grateful to Zhenwei Tang and Julian Minder for helpful discussions and feedback on this work.

\bibliographystyle{iclr2027_conference}
\bibliography{example_paper}

@inproceedings{west2026tandem,
  title={Tandem Training for Language Models},
  author={West, Robert and Anderson, Ashton and Kamar, Ece and Horvitz, Eric},
  booktitle={Proceedings of the 19th Conference of the European Chapter of the Association for Computational Linguistics (Volume 1: Long Papers)},
  pages={8265--8278},
  year={2026}
}

@article{hamade2024designing,
  title={Designing skill-compatible AI: Methodologies and frameworks in chess},
  author={Hamade, Karim and McIlroy-Young, Reid and Sen, Siddhartha and Kleinberg, Jon and Anderson, Ashton},
  journal={arXiv preprint arXiv:2405.05066},
  year={2024}
}

@article{shao2024deepseekmath,
  title={Deepseekmath: Pushing the limits of mathematical reasoning in open language models},
  author={Shao, Zhihong and Wang, Peiyi and Zhu, Qihao and Xu, Runxin and Song, Junxiao and Bi, Xiao and Zhang, Haowei and Zhang, Mingchuan and Li, YK and Wu, Yang and others},
  journal={arXiv preprint arXiv:2402.03300},
  year={2024}
}

@article{bai2022constitutional,
  title={Constitutional ai: Harmlessness from ai feedback},
  author={Bai, Yuntao and Kadavath, Saurav and Kundu, Sandipan and Askell, Amanda and Kernion, Jackson and Jones, Andy and Chen, Anna and Goldie, Anna and Mirhoseini, Azalia and McKinnon, Cameron and others},
  journal={arXiv preprint arXiv:2212.08073},
  year={2022}
}

@inproceedings{lightman2023let,
  title={Let's verify step by step},
  author={Lightman, Hunter and Kosaraju, Vineet and Burda, Yuri and Edwards, Harrison and Baker, Bowen and Lee, Teddy and Leike, Jan and Schulman, John and Sutskever, Ilya and Cobbe, Karl},
  booktitle={The twelfth international conference on learning representations},
  year={2023}
}

@phdthesis{kwon2025vllm,
  title={vLLM: An Efficient Inference Engine for Large Language Models},
  author={Kwon, Woosuk},
  year={2025},
  school={UC Berkeley}
}

@article{tan2025deepscaler,
  title={DeepScaleR: Effective RL Scaling of Reasoning Models via Iterative Context Lengthening},
  author={Tan, Sijun and Luo, Michael and Wong, Justin and Cai, Colin and Shi, Xiaoxiang and Tang, William Yuan and Roongta, Manan and Zhang, Tianjun and Li, Li Erran and Popa, Raluca Ada and others},
  year={2025}
}

@article{chen2021evaluating,
  title={Evaluating large language models trained on code},
  author={Chen, Mark and Tworek, Jerry and Jun, Heewoo and Yuan, Qiming and Pinto, Henrique Ponde De Oliveira and Kaplan, Jared and Edwards, Harri and Burda, Yuri and Joseph, Nicholas and Brockman, Greg and others},
  journal={arXiv preprint arXiv:2107.03374},
  year={2021}
}

@misc{Hendrycks2021MATHDataset,
  title         = {Measuring Mathematical Problem Solving With the {MATH} Dataset},
  author        = {Hendrycks, Dan and Burns, Collin and Kadavath, Saurav and Arora, Akul and Basart, Steven and Tang, Eric and Song, Dawn and Steinhardt, Jacob},
  year          = {2021},
  eprint        = {2103.03874},
  archivePrefix = {arXiv},
  primaryClass  = {cs.LG},
  doi           = {10.48550/arXiv.2103.03874},
  url           = {https://arxiv.org/abs/2103.03874},
  note          = {NeurIPS 2021 Datasets and Benchmarks Track}
}

@article{yang2025qwen3,
  title={Qwen3 technical report},
  author={Yang, An and Li, Anfeng and Yang, Baosong and Zhang, Beichen and Hui, Binyuan and Zheng, Bo and Yu, Bowen and Gao, Chang and Huang, Chengen and Lv, Chenxu and others},
  journal={arXiv preprint arXiv:2505.09388},
  year={2025}
}

@article{lewkowycz2022solving,
  title={Solving quantitative reasoning problems with language models},
  author={Lewkowycz, Aitor and Andreassen, Anders and Dohan, David and Dyer, Ethan and Michalewski, Henryk and Ramasesh, Vinay and Slone, Ambrose and Anil, Cem and Schlag, Imanol and Gutman-Solo, Theo and others},
  journal={Advances in neural information processing systems},
  volume={35},
  pages={3843--3857},
  year={2022}
}

@article{zheng2025group,
  title={Group sequence policy optimization},
  author={Zheng, Chujie and Liu, Shixuan and Li, Mingze and Chen, Xiong-Hui and Yu, Bowen and Gao, Chang and Dang, Kai and Liu, Yuqiong and Men, Rui and Yang, An and others},
  journal={arXiv preprint arXiv:2507.18071},
  year={2025}
}

@article{zhao2025geometric,
  title={Geometric-mean policy optimization},
  author={Zhao, Yuzhong and Liu, Yue and Liu, Junpeng and Chen, Jingye and Wu, Xun and Hao, Yaru and Lv, Tengchao and Huang, Shaohan and Cui, Lei and Ye, Qixiang and others},
  journal={arXiv preprint arXiv:2507.20673},
  year={2025}
}

@article{chu2025gpg,
  title={Gpg: A simple and strong reinforcement learning baseline for model reasoning},
  author={Chu, Xiangxiang and Huang, Hailang and Zhang, Xiao and Wei, Fei and Wang, Yong},
  journal={arXiv preprint arXiv:2504.02546},
  year={2025}
}

@article{zeng2025shrinking,
  title={Shrinking the Variance: Shrinkage Baselines for Reinforcement Learning with Verifiable Rewards},
  author={Zeng, Guanning and Zhou, Zhaoyi and Arora, Daman and Zanette, Andrea},
  journal={arXiv preprint arXiv:2511.03710},
  year={2025}
}

@article{liu2025understanding,
  title={Understanding r1-zero-like training: A critical perspective},
  author={Liu, Zichen and Chen, Changyu and Li, Wenjun and Qi, Penghui and Pang, Tianyu and Du, Chao and Lee, Wee Sun and Lin, Min},
  journal={arXiv preprint arXiv:2503.20783},
  year={2025}
}

@article{guo2025deepseek,
  title={Deepseek-r1: Incentivizing reasoning capability in llms via reinforcement learning},
  author={Guo, Daya and Yang, Dejian and Zhang, Haowei and Song, Junxiao and Wang, Peiyi and Zhu, Qihao and Xu, Runxin and Zhang, Ruoyu and Ma, Shirong and Bi, Xiao and others},
  journal={arXiv preprint arXiv:2501.12948},
  year={2025}
}

@misc{yu2025dapo,
  title         = {{DAPO}: An Open-Source {LLM} Reinforcement Learning System at Scale},
  author        = {Yu, Qiying and Zhang, Zheng and Zhu, Ruofei and Yuan, Yufeng and Zuo, Xiaochen and Yue, Yu and Dai, Weinan and Fan, Tiantian and Liu, Gaohong and others},
  year          = {2025},
  eprint        = {2503.14476},
  archivePrefix = {arXiv},
  primaryClass  = {cs.LG},
  doi           = {10.48550/arXiv.2503.14476},
  url           = {https://arxiv.org/abs/2503.14476}
}

@article{davidson2025collaboration,
  title={The Collaboration Gap},
  author={Davidson, Tim R and Fourney, Adam and Amershi, Saleema and West, Robert and Horvitz, Eric and Kamar, Ece},
  journal={arXiv preprint arXiv:2511.02687},
  year={2025}
}

@article{chu2025ssr,
  title={Ssr: Speculative parallel scaling reasoning in test-time},
  author={Chu, Yuanlin and Wang, Bo and Liu, Xiang and Chen, Hong and Liu, Aiwei and Hu, Xuming},
  journal={arXiv preprint arXiv:2505.15340},
  year={2025}
}

@article{wang2025mars,
  title={MARS: toward more efficient multi-agent collaboration for LLM reasoning},
  author={Wang, Xiao and Wang, Jia and Wang, Yijie and Dang, Pengtao and Cao, Sha and Zhang, Chi},
  journal={arXiv preprint arXiv:2509.20502},
  year={2025}
}

@article{karpov2025steganographic,
  title={The steganographic potentials of language models},
  author={Karpov, Artem and Adeleke, Tinuade and Cho, Seong Hah and Perez-Campanero, Natalia},
  journal={arXiv preprint arXiv:2505.03439},
  year={2025}
}

@article{skaf2025large,
  title={Large language models can learn and generalize steganographic chain-of-thought under process supervision},
  author={Skaf, Joey and Ibanez-Lissen, Luis and McCarthy, Robert and Watts, Connor and Georgiv, Vasil and Whittingham, Hannes and Gonzalez-Manzano, Lorena and Lindner, David and Tice, Cameron and Young, Edward James and others},
  journal={arXiv preprint arXiv:2506.01926},
  year={2025}
}

@article{burns2023weak,
  title={Weak-to-strong generalization: Eliciting strong capabilities with weak supervision},
  author={Burns, Collin and Izmailov, Pavel and Kirchner, Jan Hendrik and Baker, Bowen and Gao, Leo and Aschenbrenner, Leopold and Chen, Yining and Ecoffet, Adrien and Joglekar, Manas and Leike, Jan and others},
  journal={arXiv preprint arXiv:2312.09390},
  year={2023}
}

@inproceedings{leviathan2023fast,
  title={Fast inference from transformers via speculative decoding},
  author={Leviathan, Yaniv and Kalman, Matan and Matias, Yossi},
  booktitle={International Conference on Machine Learning},
  pages={19274--19286},
  year={2023},
  organization={PMLR}
}

@article{meng2026sparse,
  title={Sparse but critical: A token-level analysis of distributional shifts in RLVR fine-tuning of LLMs},
  author={Meng, Haoming and Huang, Kexin and Wei, Shaohang and Ma, Chiyu and Yang, Shuo and Wang, Xue and Wang, Guoyin and Ding, Bolin and Zhou, Jingren},
  journal={arXiv preprint arXiv:2603.22446},
  year={2026}
}

@inproceedings{li2025impact,
  title={The impact of language mixing on bilingual llm reasoning},
  author={Li, Yihao and Xin, Jiayi and Miao, Miranda Muqing and Long, Qi and Ungar, Lyle},
  booktitle={Proceedings of the 2025 Conference on Empirical Methods in Natural Language Processing},
  pages={32519--32536},
  year={2025}
}

@article{zhang2025improving,
  title={Improving sampling efficiency in rlvr through adaptive rollout and response reuse},
  author={Zhang, Yuheng and Yao, Wenlin and Yu, Changlong and Liu, Yao and Yin, Qingyu and Yin, Bing and Yun, Hyokun and Li, Lihong},
  journal={arXiv preprint arXiv:2509.25808},
  year={2025}
}

@article{zhan2025exgrpo,
  title={Exgrpo: Learning to reason from experience},
  author={Zhan, Runzhe and Li, Yafu and Wang, Zhi and Qu, Xiaoye and Liu, Dongrui and Shao, Jing and Wong, Derek F and Cheng, Yu},
  journal={arXiv preprint arXiv:2510.02245},
  year={2025}
}

@article{zheng2025act,
  title={Act only when it pays: Efficient reinforcement learning for llm reasoning via selective rollouts},
  author={Zheng, Haizhong and Zhou, Yang and Bartoldson, Brian R and Kailkhura, Bhavya and Lai, Fan and Zhao, Jiawei and Chen, Beidi},
  journal={arXiv preprint arXiv:2506.02177},
  year={2025}
}

@article{tang2024maia,
  title={Maia-2: A unified model for human-ai alignment in chess},
  author={Tang, Zhenwei and Jiao, Difan and McIlroy-Young, Reid and Kleinberg, Jon and Sen, Siddhartha and Anderson, Ashton},
  journal={Advances in Neural Information Processing Systems},
  volume={37},
  pages={20919--20944},
  year={2024}
}

@article{tang2026maia4all,
  title={Learning to Imitate with Less: Efficient Individual Behavior Modeling in Chess},
  author={Tang, Zhenwei and Jiao, Difan and Xue, Eric and McIlroy-Young, Reid and Kleinberg, Jon and Sen, Siddhartha and Anderson, Ashton},
  journal={Transactions on Machine Learning Research},
  issn={2835-8856},
  year={2026},
  url={https://openreview.net/forum?id=iw4kjcw319}
}

@inproceedings{magister2023teaching,
  title={Teaching small language models to reason},
  author={Magister, Lucie Charlotte and Mallinson, Jonathan and Adamek, Jakub and Malmi, Eric and Severyn, Aliaksei},
  booktitle={Proceedings of the 61st Annual Meeting of the Association for Computational Linguistics (Volume 2: Short Papers)},
  pages={1773--1781},
  year={2023}
}

@article{uesato2022solving,
  title={Solving math word problems with process-and outcome-based feedback},
  author={Uesato, Jonathan and Kushman, Nate and Kumar, Ramana and Song, Francis and Siegel, Noah and Wang, Lisa and Creswell, Antonia and Irving, Geoffrey and Higgins, Irina},
  journal={arXiv preprint arXiv:2211.14275},
  year={2022}
}

@article{kirchner2024prover,
  title={Prover-verifier games improve legibility of llm outputs},
  author={Kirchner, Jan Hendrik and Chen, Yining and Edwards, Harri and Leike, Jan and McAleese, Nat and Burda, Yuri},
  journal={arXiv preprint arXiv:2407.13692},
  year={2024}
}

@article{li2026rethinking,
  title={Rethinking on-policy distillation of large language models: Phenomenology, mechanism, and recipe},
  author={Li, Yaxuan and Zuo, Yuxin and He, Bingxiang and Zhang, Jinqian and Xiao, Chaojun and Qian, Cheng and Yu, Tianyu and Gao, Huan-ang and Yang, Wenkai and Liu, Zhiyuan and others},
  journal={arXiv preprint arXiv:2604.13016},
  year={2026}
}

@article{fu2025cache,
  title={Cache-to-Cache: Direct Semantic Communication Between Large Language Models},
  author={Fu, Tianyu and Min, Zihan and Zhang, Hanling and Yan, Jichao and Dai, Guohao and Ouyang, Wanli and Wang, Yu},
  journal={arXiv preprint arXiv:2510.03215},
  year={2025}
}

@article{zou2025latent,
  title={Latent Collaboration in Multi-Agent Systems},
  author={Zou, Jiaru and Qiu, Ruizhong and Li, Gaotang and Yang, Xiyuan and Tieu, Katherine and Lu, Pan and Shen, Ke and Tong, Hanghang and Choi, Yejin and He, Jingrui and Zou, James and Wang, Mengdi and Yang, Ling},
  journal={arXiv preprint arXiv:2511.20639},
  year={2025}
}

@article{jiao2026thinktwice,
  title={{ThinkTwice}: Jointly Optimizing Large Language Models for Reasoning and Self-Refinement},
  author={Jiao, Difan and Wen, Qianfeng and Yang, Blair and Tang, Zhenwei and Anderson, Ashton},
  journal={arXiv preprint arXiv:2604.01591},
  year={2026}
}

@article{roytburg2026measuring,
  title={Measuring Weak-to-Strong Legibility of Reasoning Models},
  author={Roytburg, Dani and Sridhar, Shreya and Ippolito, Daphne},
  journal={arXiv preprint arXiv:2603.20508},
  year={2026}
}

@misc{arcprize2026astra,
  author={{ARC Prize Foundation}},
  title={{OpenAI}'s {GPT-6 Astra} on {ARC-AGI-3}},
  year={2026},
  howpublished={\url{https://arcprize.org/blog/astra}}
}

@misc{pachocki2026alien,
  author={Pachocki, Jakub},
  title={An Alien Mind},
  year={2026},
  howpublished={\url{https://openai.com/index/an-alien-mind/}}
}

@article{jose2025illegible,
  author={Jose, Arun},
  title={Reasoning Models Sometimes Output Illegible Chains of Thought},
  journal={arXiv preprint arXiv:2510.27338},
  year={2025}
}

@book{nasem2022humanai,
  author={{National Academies of Sciences, Engineering, and Medicine}},
  title={Human-{AI} Teaming: State-of-the-Art and Research Needs},
  publisher={The National Academies Press},
  address={Washington, DC},
  year={2022},
  doi={10.17226/26355}
}

@article{shin2023superhuman,
  author={Shin, Minkyu and Kim, Jin and van Opheusden, Bas and Griffiths, Thomas L},
  title={Superhuman artificial intelligence can improve human decision-making by increasing novelty},
  journal={Proceedings of the National Academy of Sciences},
  volume={120},
  number={12},
  pages={e2214840120},
  year={2023}
}

@article{panfilov2026stealing,
  author={Panfilov, Alexander and Schmotz, David and Shumailov, Ilia and Beurer-Kellner, Luca and Schaeffer, Joachim and Prabhu, Ameya and Geiping, Jonas and Andriushchenko, Maksym},
  title={Stealing Reasoning Traces from Proprietary {LLM} {APIs}},
  journal={arXiv preprint arXiv:2608.09867},
  year={2026}
}

@article{teoh2017rage,
  author={Teoh, Eric R and Kidd, David G},
  title={Rage against the machine? {Google}'s self-driving cars versus human drivers},
  journal={Journal of Safety Research},
  volume={63},
  pages={57--60},
  year={2017}
}

@inproceedings{he2024olympiadbench,
  title     = {{OlympiadBench}: A Challenging Benchmark for Promoting {AGI} with Olympiad-Level Bilingual Multimodal Scientific Problems},
  author    = {He, Chaoqun and Luo, Renjie and Bai, Yuzhuo and Hu, Shengding and Thai, Zhen Leng and Shen, Junhao and Hu, Jinyi and Han, Xu and Huang, Yujie and Zhang, Yuxiang and Liu, Jie and Qi, Lei and Liu, Zhiyuan and Sun, Maosong},
  booktitle = {Proceedings of the 62nd Annual Meeting of the Association for Computational Linguistics (Volume 1: Long Papers)},
  pages     = {3828--3850},
  year      = {2024}
}

@article{wynn2025talk,
  title={Talk Isn't Always Cheap: Understanding Failure Modes in Multi-Agent Debate},
  author={Wynn, Andrea and Satija, Harsh and Hadfield, Gillian},
  journal={arXiv preprint arXiv:2509.05396},
  year={2025},
  url={https://arxiv.org/abs/2509.05396}
}

@article{li2025small,
  title={Small Models Struggle to Learn from Strong Reasoners},
  author={Li, Yuetai and Yue, Xiang and Xu, Zhangchen and Jiang, Fengqing and Niu, Luyao and Lin, Bill Yuchen and Ramasubramanian, Bhaskar and Poovendran, Radha},
  journal={arXiv preprint arXiv:2502.12143},
  year={2025},
  url={https://arxiv.org/abs/2502.12143}
}

@article{baker2025monitoring,
  title={Monitoring Reasoning Models for Misbehavior and the Risks of Promoting Obfuscation},
  author={Baker, Bowen and Huizinga, Joost and Gao, Leo and Dou, Zehao and Guan, Melody Y. and Madry, Aleksander and Zaremba, Wojciech and Pachocki, Jakub and Farhi, David},
  journal={arXiv preprint arXiv:2503.11926},
  year={2025},
  url={https://arxiv.org/abs/2503.11926}
}

@inproceedings{drori2026output,
  title={Output Supervision Can Obfuscate the Chain of Thought},
  author={Drori, Jacob and Marks, Luke and Woodworth, Bryce and Cloud, Alex and Turner, Alexander},
  booktitle={International Conference on Learning Representations},
  year={2026},
  url={https://proceedings.iclr.cc/paper_files/paper/2026/hash/53813b70e97b6e5778e24fd88e1259f1-Abstract-Conference.html}
}

@article{cetin2025teachers,
  title={Reinforcement Learning Teachers of Test Time Scaling},
  author={Cetin, Edoardo and Zhao, Tianyu and Tang, Yujin},
  journal={arXiv preprint arXiv:2506.08388},
  year={2025},
  url={https://arxiv.org/abs/2506.08388}
}

@article{anwar2026monitorability,
  title={Analyzing and Improving Chain-of-Thought Monitorability Through Information Theory},
  author={Anwar, Usman and Bakker, Tim and Kianfar, Dana and Pinneri, Cristina and Louizos, Christos},
  journal={arXiv preprint arXiv:2602.18297},
  year={2026},
  url={https://arxiv.org/abs/2602.18297}
}

@article{surana2026rollout,
  title={Generate, Filter, Control, Replay: A Comprehensive Survey of Rollout Strategies for {LLM} Reinforcement Learning},
  author={Surana, Rohan and Mundada, Gagan and Jiang, Xunyi and Wang, Chuhan and Tang, Zhenwei and Jiao, Difan and Huang, Zihan and Xiong, Yuxin and Wu, Junda and Yu, Sheldon and Li, Xintong and Jain, Raghav and Kuang, Nikki and Zhou, Sizhe and Jin, Bowen and Chu, Zhendong and Yu, Tong and Rossi, Ryan and Huang, Kuan-Hao and Shang, Jingbo and Han, Jiawei and McAuley, Julian},
  journal={arXiv preprint arXiv:2605.02913},
  year={2026},
  url={https://arxiv.org/abs/2605.02913}
}

@inproceedings{jiao2026siren,
  title={{LLM} Safety From Within: Detecting Harmful Content with Internal Representations},
  author={Jiao, Difan and Liu, Yilun and Yuan, Ye and Tang, Zhenwei and Du, Linfeng and Wu, Haolun and Anderson, Ashton},
  booktitle={Proceedings of the 64th Annual Meeting of the Association for Computational Linguistics (Volume 1: Long Papers)},
  year={2026},
  publisher={Association for Computational Linguistics},
  pages={39711--39727},
  doi={10.18653/v1/2026.acl-long.1844},
  url={https://aclanthology.org/2026.acl-long.1844/}
}

@inproceedings{jiao2024spin,
  title={{SPIN}: Sparsifying and Integrating Internal Neurons in Large Language Models for Text Classification},
  author={Jiao, Difan and Liu, Yilun and Tang, Zhenwei and Matter, Daniel and Pfeffer, J{\"u}rgen and Anderson, Ashton},
  booktitle={Findings of the Association for Computational Linguistics: ACL 2024},
  year={2024},
  publisher={Association for Computational Linguistics},
  pages={4666--4682},
  doi={10.18653/v1/2024.findings-acl.277},
  url={https://aclanthology.org/2024.findings-acl.277/}
}
\clearpage
\appendix
\section{Extended Related Work}
\label{app:extended-related}

\xhdr{Reinforcement learning with verifiable rewards} Recent advances in reasoning post-training have leveraged reinforcement learning with verifiable rewards (RLVR). \citet{shao2024deepseekmath} introduced Group Relative Policy Optimization (GRPO) as a practical recipe for this setting, and \citet{guo2025deepseek} demonstrated that outcome-only RL can elicit strong reasoning capabilities without process supervision. A growing follow-up literature expands the GRPO design space along two axes: optimization-side variants revisit stability, optimization biases, and ratio aggregation~\citep{yu2025dapo,liu2025understanding,zheng2025group,zhao2025geometric,chu2025gpg,zeng2025shrinking,surana2026rollout}, while rollout-side work targets sampling efficiency, prompt filtering, and reuse of high-signal trajectories~\citep{zhang2025improving,zhan2025exgrpo,zheng2025act}. RLVR has also been extended beyond one-shot solution generation to jointly optimize reasoning and self-refinement under the same verifiable outcome reward~\citep{jiao2026thinktwice}. \modelname{} is complementary to these directions: it changes which model emits each token within a single rollout while leaving the task reward, verifier, GRPO loss, and rollout-management strategy unchanged.

\xhdr{Multi-model reasoning and transfer} Drafter--target and role-based multi-agent systems divide inference across models to improve test-time efficiency or reasoning quality~\citep{chu2025ssr,wang2025mars}. Internal representations can also support inter-model communication and knowledge extraction through hidden states and KV caches, bypassing explicit textual handoffs~\citep{jiao2024spin, fu2025cache,zou2025latent,jiao2026siren}. These works study an alternative communication interface for multi-model reasoning. Separately, successful on-policy distillation (OPD) depends on compatibility between teacher and student token distributions~\citep{li2026rethinking}. \modelname{} pursues a complementary direction: it retains the ordinary token interface and places the frozen weaker partner inside the reward-bearing rollout, so that compatibility is learned as part of the senior's reasoning post-training.

\section{Reproducibility}

\subsection{Tandem rollout implementation}
\label{app:rollout}

A naive realization of tandem rollout couples two HuggingFace models in an outer Python loop with manual KV-cache management. We built such a prototype and found it impractical for RL training: at 512 generated tokens it exhausts the memory of a single 80~GB GPU, making RLVR under long chain-of-thoughts for mathematical reasoning tasks infeasible. Moreover, per-step throughput at shorter lengths is already roughly $4\times$ slower than a single-model vLLM \citep{kwon2025vllm} rollout. Since RL training is dominated by rollout time, this slowdown alone makes large-scale tandem RL infeasible.

We therefore implement the tandem rollout \emph{inside} vLLM, with both $\pi_{\text{sen}}$ and $\pi_{\text{jun}}$ treated as paged-attention-aware models in the same engine. The result is a decoding path that pays only the dual-forward cost (theoretically twice the single-model latency) and inherits every other vLLM optimization: paged KV cache, continuous batching, FlashAttention, CUDA graphs, and tensor parallelism. Empirically our throughput is approximately $0.5\times$ that of a single-model vLLM rollout, matching the architectural lower bound for sequential dual decoding.

\subsubsection{Dual-decoder backend}
\label{app:rollout-backend}

\begin{figure}[t]
  \centering
  \includegraphics[width=\linewidth]{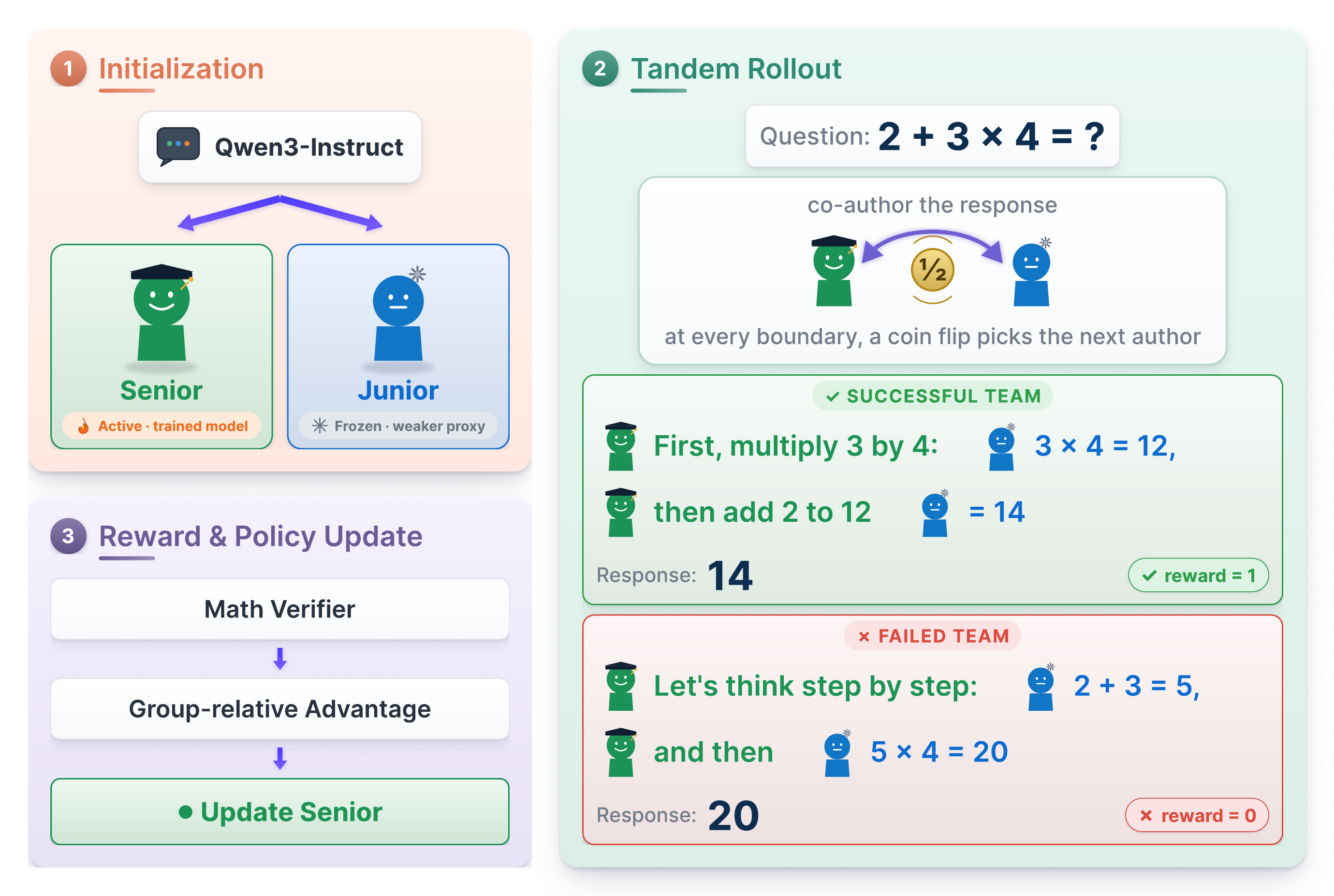}
  \caption{\modelname{} at a glance.}
  \label{fig:glance}
\end{figure}

In our implementation, $\pi_{\text{sen}}$ and $\pi_{\text{jun}}$ are loaded onto separate devices and registered in the same vLLM engine, with the junior's layers carrying a fixed name prefix so that the engine's forward context can route them without interfering with the senior's. The two models share per-step attention metadata (slot mappings, block tables, position indices), since they process identical token sequences, but each writes to its own KV-cache tensor on its own device. At each generation step the engine performs one forward pass through $\pi_{\text{sen}}$, then one through $\pi_{\text{jun}}$ on the same appended token, samples independently from each model's logits, and a tandem sampler emits the chosen token according to the active-model schedule of \S\ref{sec:rollout}; the chosen token is then fed to both models on the next step, keeping the two KV caches in lockstep with the shared response $y_{<t}$. This isolates the tandem-specific logic to model loading, the second forward, and the sampler, while leaving every other vLLM runtime path untouched.

For each request, the engine also accumulates a per-token authorship stream alongside the generated tokens, flagging which positions were emitted by $\pi_{\text{sen}}$. The stream is propagated through the same output channels as the tokens themselves (model-runner output, engine-core output, per-request completion output) and is exposed to the verl trainer as a tensor field of the rollout batch. On the trainer side, the actor's loss path consumes it by elementwise-multiplying it into the per-token response mask before the policy-gradient sum, which yields exactly the senior-only loss of \S\ref{sec:loss}. This is the only addition to the verl GRPO surface; the actor, critic, and trainer code paths are otherwise unchanged.

\subsubsection{Word-level handoff details}
\label{app:rollout-word}

With the vLLM-native backend in place, we now detail the word-level handoff schedule of \S\ref{sec:rollout} as instantiated in our implementation.

The set of word-boundary tokens is auto-resolved from the active tokenizer at engine initialization. For any BPE tokenizer using a leading-space marker (the Qwen3 / GPT-style word-initial prefix), The boundary set is exactly the set of vocabulary IDs whose surface form begins with that marker. For the Qwen3 tokenizer used in this work, this yields roughly $53\text{k}$ of $151\text{k}$ IDs.

The boundary set alone, however, is not sufficient. Mathematical reasoning produces long runs of non-boundary tokens inside internal LaTeX expressions, variable identifiers, and code blocks, where leading-space tokens never appear. Without intervention, such runs would receive no handoff, allowing the active model to retain control across an entire derivation step and reintroducing exactly the reward-hacking surface that word-level granularity was meant to close. We therefore add a backstop: once $K$ consecutive non-boundary tokens have elapsed since the last redraw, the active model is redrawn under the same Bernoulli($p$) rule used at boundaries. We use $K = 32$, chosen as a default that prevented monopoly behaviour in our early rollouts.

The Bernoulli draws at boundary and cap-fallback positions use the per-request random number generator when available and otherwise use PyTorch's global generator. To avoid replaying the full token history at every generation step, the tandem sampler maintains a small per-request state (next active model, tokens since the last redraw) and resumes from this state on each subsequent step.

\begin{table}[t]
\centering
\caption{Hyperparameters for training \modelname{} on DeepScaleR with Qwen3-4B-Instruct-2507. The GRPO training block matches the vanilla GRPO baseline; the tandem-rollout block lists the additional \modelname{} settings.}
\begin{tabular}{@{}ll@{}}
\toprule
\textbf{Category} & \textbf{Hyperparameter} \\
\midrule
\multicolumn{2}{l}{\textit{GRPO training}} \\
\midrule
Learning rate & $1 \times 10^{-6}$ \\
PPO clip ratio & $0.2$ \\
Train batch size & $16$ \\
PPO mini-batch size & $8$ \\
Group size $G$ (rollouts per prompt) & $8$ \\
Max response length & $3000$ \\
Entropy coefficient & $0$ \\
Advantage estimator & GRPO (group-relative, std-normalised) \\
\midrule
\multicolumn{2}{l}{\textit{Tandem rollout (\modelname{}-specific)}} \\
\midrule
Selection strategy & word-level \\
Handoff probability $p$ & $0.5$ \\
Subword-span cap $K$ & $32$ \\
Junior-token loss weight $\lambda_{\text{jun}}$ & $0$ \\
Senior $\pi_{\text{sen}}$ initialisation & Qwen3-4B-Instruct \\
Junior $\pi_{\text{jun}}$ initialisation & Qwen3-4B-Instruct \\
\midrule
\multicolumn{2}{l}{\textit{Generation}} \\
\midrule
Train temperature & $0.6$ \\
Train top-$p$ & $1.0$ \\
Train top-$k$ & $-1$ \\
Evaluation temperature & $0.7$ \\
Evaluation top-$p$ & $0.8$ \\
Evaluation top-$k$ & $20$ \\
\bottomrule
\end{tabular}
\label{tab:hparams}
\end{table}

\subsection{Datasets}
\label{app:datasets}

We train on DeepScaleR~\citep{tan2025deepscaler}, a corpus of approximately $40{,}000$ competition-math problems compiled from past olympiad and contest sources. Each example provides a problem statement and a verifiable final answer; reward is computed by extracting the final \texttt{\textbackslash boxed\{\}} expression from the model's response and comparing it against the gold answer with a sympy-based numeric or symbolic match adopted from \citet{Hendrycks2021MATHDataset}.
For reproduction, the released data-preparation script holds out $1{,}000$ DeepScaleR problems from training for validation.

For evaluation we use a panel of four competition-math benchmarks, each disjoint from the training corpus:
\begin{itemize}
    \item \textbf{AMC 23--25}~\footnote{Compiled from AMC 10/12 problems (American Mathematics Competitions) of 2023, 2024, and 2025.}: 121 problems.
    \item \textbf{AIME 24--26}~\footnote{Compiled from AIME (American Invitational Mathematics Examination) of 2024, 2025, and 2026.}: 90 problems.
    \item \textbf{Minerva Math}~\footnote{\url{https://huggingface.co/datasets/math-ai/minervamath}}: 272 problems.
    \item \textbf{OlympiadBench}\footnote{\url{https://huggingface.co/datasets/Hothan/OlympiadBench}}: 581 olympiad-level problems.
\end{itemize}

Each problem is formatted with an instruction requiring the model to output its final answer in \texttt{\textbackslash boxed\{\}} form. We score every rollout with the same boxed-extraction grader used during training; aggregate metrics are reported as pass@$k$ over $k$ independent rollouts per problem following the unbiased estimator proposed in \citet{chen2021evaluating}.

\subsection{Hyperparameter configuration}
\label{app:hparams}

Table~\ref{tab:hparams} summarises the hyperparameters used to train \modelname{} and the vanilla GRPO baseline. All \modelname{} experiments reported in this paper were trained on 2$\times$ NVIDIA A100 80GB GPUs, with $\pi_{\text{sen}}$ on \texttt{cuda:0} and $\pi_{\text{jun}}$ on \texttt{cuda:1}; both models run with tensor-parallel size $1$.

\subsection{End-to-end \modelname{} pipeline}
\label{app:algorithm}

\begin{algorithm}[H]
\caption{\modelname{}.}
\label{alg:trl}
\begin{algorithmic}[1]
\REQUIRE Base model $\pi_0$;\; training set $\mathcal{D}$;\; batch size $B$;\; group size $G$;\; max response length $L$;\; word-boundary set $\mathcal{B}$;\; subword-span cap $K$;\; Bernoulli probability $p$;\; verifier $r(\cdot)$
\STATE Initialize $\pi_{\text{sen}} \leftarrow \pi_0$ and $\pi_{\text{jun}} \leftarrow \pi_0$;\; freeze $\pi_{\text{jun}}$
\FOR{$\tau = 1, 2, \ldots$}
    \STATE Sample a batch of prompts $\{x_j\}_{j=1}^{B} \sim \mathcal{D}$
    \FOR{each $(j, i)$ with $j \in [B]$ and $i \in [G]$ }
        \STATE \textcolor{gray}{// Tandem rollout (\S\ref{sec:rollout})}
        \STATE Draw $a_1 \sim \mathrm{Bernoulli}(p)$ over $\{\text{sen}, \text{jun}\}$;\; $c \leftarrow 0$
        \FOR{$t = 1, \ldots, L$}
            \STATE Forward $\pi_{\text{sen}}$ and $\pi_{\text{jun}}$ on $x_j \cdot y_{<t}$;\; sample $y_t^{\text{sen}}, y_t^{\text{jun}}$ independently
            \STATE $y_t \leftarrow y_t^{a_t}$;\quad $m_t \leftarrow \mathds{1}[a_t = \text{sen}]$
            \STATE \textbf{break} if $y_t$ is the end-of-sequence token
            \IF{$y_t \in \mathcal{B}$ \textbf{or} $c \geq K$}
                \STATE Draw $a_{t+1} \sim \mathrm{Bernoulli}(p)$ over $\{\text{sen}, \text{jun}\}$;\; $c \leftarrow 0$
            \ELSE
                \STATE $a_{t+1} \leftarrow a_t$;\; $c \leftarrow c + 1$
            \ENDIF
        \ENDFOR
        \STATE Record $y^{(j,i)} \leftarrow y_{1:t}$ and authorship mask $m^{(j,i)} \leftarrow m_{1:t}$
    \ENDFOR
    \STATE \textcolor{gray}{// Senior-only GRPO update (\S\ref{sec:loss})}
    \STATE Compute rewards $r^{(j,i)} \leftarrow r(y^{(j,i)})$
    \STATE Compute group-relative advantages $A^{(j,i)}_t$ from $\{r^{(j,i)}\}_{i=1}^{G}$
    \STATE Update $\pi_{\text{sen}}$ with the GRPO objective of \S\ref{sec:loss}, with response mask narrowed by $m^{(j,i)}$
\ENDFOR
\end{algorithmic}
\end{algorithm}

\section{Additional Results}

\subsection{Statistical significance of \modelname{} gains}
\label{app:significance}

We complement the bootstrap standard errors of \S\ref{sec:results} with paired one-sided $t$-tests on per-problem statistics, mean-pooled across AMC, AIME, and Minerva. KL-Reg here denotes the default $\beta=10^{-3}$ arm. For pass@$k$ we test per-problem unbiased estimates~\citep{chen2021evaluating}; for per-token cross-entropy and distribution overlap $\alpha$ we average per-problem measurements before testing. We note that each test is one-sided in the claimed direction, and asterisks denote significance levels: $^{*}p<0.05$, $^{**}p<0.01$, $^{***}p<0.001$.

\begin{table}[h]
\centering
\setlength{\tabcolsep}{10pt}
\caption{Paired $t$-tests for the evaluation gains of \modelname{} over baselines.}
\label{tab:significance}
\begin{tabular}{l r r l}
\toprule
Comparison & $\Delta$ & $t$ & $p$ \\
\midrule
\multicolumn{4}{l}{\textit{Handoff Accuracy} (pass@$k$, $\Delta$ in percentage points, $\uparrow$)} \\
\modelname{} $>$ GRPO,\;\; $k{=}1$  & $+1.42$  & $+2.61$  & $0.005^{**}$ \\
\modelname{} $>$ GRPO,\;\; $k{=}2$  & $+1.72$  & $+2.66$  & $0.004^{**}$ \\
\modelname{} $>$ GRPO,\;\; $k{=}4$  & $+1.88$  & $+2.32$  & $0.010^{*}$ \\
\modelname{} $>$ GRPO,\;\; $k{=}8$  & $+2.07$  & $+1.83$  & $0.034^{*}$ \\
\modelname{} $>$ KL-Reg,   $k{=}1$  & $+1.01$  & $+1.73$  & $0.043^{*}$ \\
\modelname{} $>$ KL-Reg,   $k{=}2$  & $+1.24$  & $+1.90$  & $0.029^{*}$ \\
\modelname{} $>$ KL-Reg,   $k{=}4$  & $+1.62$  & $+2.13$  & $0.017^{*}$ \\
\modelname{} $>$ KL-Reg,   $k{=}8$  & $+2.07$  & $+2.14$  & $0.016^{*}$ \\
\midrule
\multicolumn{4}{l}{\textit{Junior Legibility} (per-token cross-entropy, $\Delta$ in nats, $\downarrow$)} \\
\modelname{} $<$ GRPO    & $-0.0146$ & $\phantom{0}-9.73$ & ${<}10^{-20}\,^{***}$ \\
\modelname{} $<$ KL-Reg  & $-0.0114$ & $-10.94$            & ${<}10^{-24}\,^{***}$ \\
\midrule
\multicolumn{4}{l}{\textit{Distribution Overlap} ($\alpha$, $\Delta$ in percentage points, $\uparrow$)} \\
\modelname{} $>$ GRPO    & $+1.47$ & $+36.16$ & ${<}10^{-139}\,^{***}$ \\
\modelname{} $>$ KL-Reg  & $+1.10$ & $+30.22$ & ${<}10^{-113}\,^{***}$ \\
\bottomrule
\end{tabular}
\end{table}

Table~\ref{tab:significance} reports paired one-sided $t$-tests for the three directional claims involving \modelname{} in the main text. On handoff accuracy (\S\ref{sec:eval-handoff}), \modelname{}'s lead over both GRPO and KL-Reg holds at every $k$, with all eight tests significant at $p<0.05$ and two tests at $p<0.01$; the consistency across $k$ rules out a single-$k$ artifact. On junior legibility (\S\ref{sec:eval-legibility}), \modelname{}'s per-token cross-entropy is significantly lower than both baselines, confirming that the senior's chain-of-thought is genuinely more predictable to the junior. The distribution-overlap gap is the sharpest: \modelname{} beats both baselines at $p<10^{-100}$.

\subsection{Phrase-level view of the over-emitted vocabulary}
\label{app:token-drift}

\begin{figure}[h]
  \centering
  \includegraphics[width=0.92\linewidth]{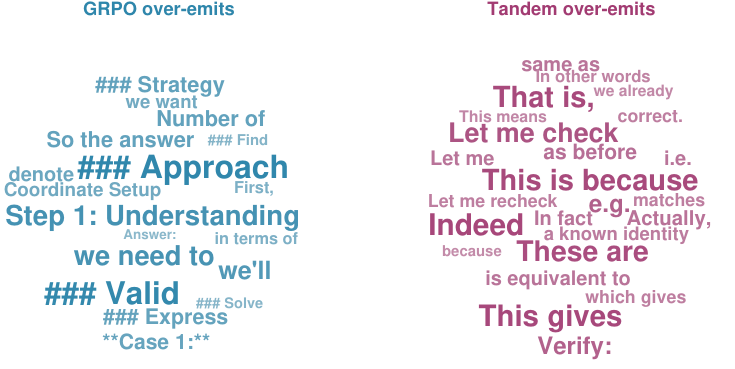}
  \caption{Surface phrases each senior over-emits relative to the other, sized by the rate ratio between the two seniors.}
  \label{fig:phrase_cloud}
\end{figure}

Figure~\ref{fig:phrase_cloud} extends the case study of \S\ref{sec:eval-drift} from tokens to surface phrases, counted per million tokens over the same generations. Each side shows phrases one senior emits at a higher rate than the other, sized by that ratio on a scale shared across the two sides. The reading matches panel (d) of Figure~\ref{fig:drift_and_distance}: GRPO's side is the furniture of a formatted report, while \modelname{}'s side is the connective and checking language that carries one step into the next. Several phrases on the \modelname{} side, such as \texttt{That is,} and \texttt{This gives}, are ones GRPO suppresses below the junior's rate while \modelname{} keeps them near it, panel (c)'s pull toward the junior in words.

\subsection{A step-level case study of junior legibility}
\label{app:step-semantics}

The aggregate results in \S\ref{sec:eval-legibility} show that the junior finds \modelname{} reasoning less surprising. We next ask whether this advantage is uniform across reasoning steps or concentrated at transitions that rely on prior context. We collect four solo generations per problem from the junior, GRPO senior, and \modelname{} senior on 211 AMC and AIME problems, for 2,532 generations in total. A reasoning step is a span between the double-newline boundaries used in our handoff evaluation. We uniformly sample up to 12 steps from each chain so that long responses do not dominate.

\xhdr{Defining the split} We call a step \emph{self-contained} if, read in isolation, it states what it is doing and makes the symbols and references needed to interpret it identifiable. A step is \emph{context-dependent} if its role or referents require earlier text, for example because it continues a prior calculation or relies on an unresolved pronoun or symbol. To apply this definition, a frozen Qwen3.6-35B judge sees only the current step, without the problem or preceding chain of thought, and is constrained to answer yes or no. We map yes to self-contained and no to context-dependent. The judge prompt is fixed throughout, and the judge is not told which model produced the step.

This isolation is deliberate: it tests whether the step itself carries enough information to be interpreted, rather than whether external context can make it interpretable. Showing the problem or earlier reasoning could supply precisely the information whose absence the split is intended to detect. This is not meant to reproduce the junior's information at inference; we restore the full problem and reasoning prefix when measuring junior surprisal below.

\vspace{1mm}
\noindent\colorbox{gray!10}{\parbox{\dimexpr\linewidth-2\fboxsep\relax}{
\small\textbf{An example pair from one evaluated \modelname{} generation.}\\[0.6mm]
\textbf{Self-contained:} ``Each game has one winner and one loser, so the total number of wins is equal to the total number of games.''\\[0.6mm]
\textbf{Context-dependent:} ``So $L=3$ works.''\\[0.6mm]
The first states both the premise and its implication locally. The second requires earlier text to identify what $L$ denotes, which condition is being tested, and why $3$ satisfies it.
}}
\vspace{1mm}

As a model-free robustness check, we also construct a regex split. After trimming whitespace, a step is marked context-dependent if its first word is one of ten fixed context-carrying terms or its final character is a colon; all other steps are marked self-contained.\footnote{The case-insensitive word list is \texttt{so}, \texttt{then}, \texttt{thus}, \texttt{this}, \texttt{it}, \texttt{therefore}, \texttt{now}, \texttt{and}, \texttt{which}, and \texttt{substituting}.}

\xhdr{Comparison} Within each group, we compare how readily the frozen junior predicts the opening of a senior's step, conditioned on the original problem and all preceding reasoning. Table~\ref{tab:step-context} reports the difference in junior cross-entropy, $\Delta\mathrm{CE} = \mathrm{CE}_{\modelname{}} - \mathrm{CE}_{\mathrm{GRPO}}$, so negative values indicate lower junior surprisal under \modelname{}. Crucially, neither split has access to these likelihoods: \emph{context-dependent} describes reliance on prior text, not whether the junior finds the step difficult.

\begin{table}[h]
\centering
\setlength{\tabcolsep}{7pt}
\caption{Context-dependent step frequencies and the \modelname{}--GRPO gap in junior cross-entropy at step entry under the LLM judge and regex splits.}
\label{tab:step-context}
\small
\begin{tabular}{lccc cc}
\toprule
\multirow{2}{*}{Split} & \multicolumn{3}{c}{Context-dependent steps (\%)} & \multicolumn{2}{c}{$\Delta\mathrm{CE}$ (nats, $\downarrow$)} \\
\cmidrule(lr){2-4}\cmidrule(lr){5-6}
& Junior & GRPO & \modelname{} & Self-contained & Context-dependent \\
\midrule
LLM judge & 37.6 & 35.5 & 33.5 & $+0.0024$ & $-0.0707$ \\
Regex split & 41.7 & 39.3 & 37.1 & $+0.0055$ & $-0.0667$ \\
\bottomrule
\end{tabular}
\end{table}

\xhdr{Results} The self-contained groups exhibit near-zero \modelname{}--GRPO gaps under both splits. By contrast, the context-dependent groups show reductions of $0.067$--$0.071$ nats in junior cross-entropy for \modelname{}. The splits agree on only $73.5\%$ of steps, yet recover nearly identical conditional gaps. \modelname{} also produces slightly fewer context-dependent steps, but prevalence alone does not explain the difference observed within that group.

Because self-containedness is an operational distinction and the semantic split is determined by an LLM judge rather than human annotation, we treat this analysis as a descriptive case study, not a causal account or a universal partition of reasoning steps. Across both splits, the observed junior-legibility difference is concentrated at transitions whose interpretation leans on prior context.

\subsection{Controls for junior cross-entropy}
\label{app:ce-controls}

We use two controls to test whether junior cross-entropy is related to task success and whether its reduction under \modelname{} can be explained by the predictability of \modelname{} traces under the senior itself.

\subsubsection{Relation to task success}

We first test whether lower junior cross-entropy is associated with a trace's correctness. We pair every trace used in Table~\ref{tab:legibility} with its binary correctness reward. For each senior, we retain problems containing both correct and incorrect traces among the eight solo rollout samples and compute the junior cross-entropy of correct traces minus that of incorrect traces within each problem. Negative values mean successful traces are less surprising to the junior. We average these differences over problems and obtain 95\% confidence intervals by bootstrapping problems.

\begin{table}[h]
\centering
\setlength{\tabcolsep}{10pt}
\caption{Difference in junior cross-entropy between correct and incorrect senior traces within the same problem (nats, $\downarrow$).}
\label{tab:surprisal-success}
\begin{tabular}{lccc}
\toprule
Senior & Mixed problems & $\Delta\mathrm{CE}$ & 95\% CI \\
\midrule
GRPO & 239 & $-0.0027$ & $[-0.0060,\,+0.0005]$ \\
\modelname{} & 242 & $-0.0078$ & $[-0.0112,\,-0.0044]$ \\
\bottomrule
\end{tabular}
\end{table}

Pooling the two seniors, the within-problem difference between correct and incorrect traces is $-0.0053$ nats, with a 95\% confidence interval of $[-0.0079,\,-0.0026]$. The point estimate is more negative for \modelname{}, and only its interval excludes zero. Thus lower junior surprisal is associated with task success on these solo traces. This comparison does not identify a causal mechanism or establish semantic understanding.

\subsubsection{Cross-entropy under the senior}

We next test whether \modelname{} traces are simply easier to predict under the model that generated them. We score each trace under its senior and report the difference
$\mathrm{CE}_{\mathrm{jun}}-\mathrm{CE}_{\mathrm{sen}}$ on the same observed tokens. If general predictability explained the lower junior cross-entropy, \modelname{} would also have lower senior cross-entropy and its advantage would shrink after this subtraction.

\begin{table}[h]
\centering
\setlength{\tabcolsep}{10pt}
\caption{Junior and senior cross-entropy and their difference on senior traces, reported as a macro average across four benchmarks (nats).}
\label{tab:senior-self-ce}
\begin{tabular}{lccc}
\toprule
Senior & Junior CE & Senior CE & $\mathrm{CE}_{\mathrm{jun}}-\mathrm{CE}_{\mathrm{sen}}$ \\
\midrule
GRPO & $0.1294$ & $0.0636$ & $0.0658$ \\
\modelname{} & $0.1183$ & $0.0882$ & $0.0300$ \\
\bottomrule
\end{tabular}
\end{table}

The \modelname{} traces have $0.0247$ nats higher senior cross-entropy than the GRPO traces, with a 95\% confidence interval of $[0.0233,\,0.0261]$. The gap between junior and senior cross-entropy is nevertheless $0.0358$ nats lower, with a 95\% confidence interval of $[-0.0373,\,-0.0344]$. This is a $54\%$ reduction in the macro average, and the gap is smaller on all four benchmarks. Under the same decoding settings, this pattern is consistent with GRPO concentrating probability on a narrower set of continuations as it drifts from the junior, whereas \modelname{} retains more uncertainty and stays closer to the junior. Thus \modelname{}'s lower junior cross-entropy is not explained by greater predictability under the generating senior. Because the traces were sampled with temperature and top-$p$, the gap is a likelihood contrast on sampled traces rather than an exact KL divergence.

\subsection{Stochastic reasoning-step handoff accuracy}
\label{app:stoch-hr}

The main HR result in \S\ref{sec:eval-handoff} alternates the senior and the junior strictly at every \texttt{\textbackslash n\textbackslash n} boundary. This deterministic round-robin is a clean evaluation protocol, but in scenarios such as agent-to-agent collaboration, drafter-target inference, and human-AI handoffs, the timing of handoffs is normally more flexible and the next contributor is typically chosen by the situation rather than by a fixed cadence. As a robustness check, we replace the deterministic round-robin with an independent Bernoulli($p=0.5$) draw at every \texttt{\textbackslash n\textbackslash n} boundary, keeping every other setting identical. The expected senior share remains $0.5$, but the per-rollout schedule now varies, and a team's success has to be robust to that variation.

\begin{table}[h]
\centering
\setlength{\tabcolsep}{10pt}
\caption{Reasoning-step handoff accuracy under stochastic Bernoulli ($p=0.5$) schedules at \texttt{\textbackslash n\textbackslash n} boundaries (pass@$8$, \%, $\uparrow$). Junior is the Qwen3-4B-Instruct training partner.}
\label{tab:stoch_hr}
\begin{tabular}{l ccc}
\toprule
Benchmark & GRPO & \modelname{} & $\Delta$ \\
\midrule
AMC       & 78.5 & 80.2 & $+1.7$ \\
AIME      & 38.9 & 42.2 & $+3.3$ \\
Minerva   & 44.5 & 44.9 & $+0.4$ \\
Olympiad  & 64.0 & 65.1 & $+1.1$ \\
\midrule
Macro Avg & 56.5 & 58.1 & $+1.6$ \\
\bottomrule
\end{tabular}
\end{table}

Per-rollout senior shares are well-controlled across methods (mean $50.0\%$ for both, standard deviation in the $5\%$--$9\%$ range across benchmarks). \modelname{} retains a directional advantage over vanilla GRPO on every benchmark and on macro, with the largest gap again on AIME ($+3.3$ at pass@$8$). The macro gap is smaller than under the strict schedule of \S\ref{sec:eval-handoff} ($+1.6$ vs.\ $+3.1$), as expected: under a randomised schedule, individual rollouts vary in how much of the trajectory each model carries, and pass@$8$ aggregates over schedule variation in addition to the usual generation variation. The strict-alternation protocol of the main text is the more stringent measurement; the stochastic protocol here verifies that \modelname{}'s compatibility lift is not an artefact of a fixed schedule.

\section{Different training signals in KL regularization and tandem rollouts}
\label{app:ablation}

A natural alternative to \modelname{} is to keep solo GRPO rollouts and add a per-token KL penalty toward the frozen junior $\pi_{\text{jun}}$. This likewise ``anchors'' the senior toward the junior, but as a soft additive regularizer on the senior's own trajectories rather than by sampling rollouts from a tandem mixture. Although both objectives involve the same frozen junior, they expose the senior to different training signals. We make this distinction explicit below and relate it to the empirical pattern in our ablation (\S\ref{sec:ablation}).

\subsection{Setup and what each objective optimizes}
\label{app:math-setup}

Let $\mathcal{V}$ be the vocabulary, $x$ a prompt, $y = y_{1:T} \in \mathcal{V}^T$ a response, $h_t = x \cdot y_{<t}$ the context at position $t$, and $r : \mathcal{V}^T \to [0,1]$ a bounded verifier reward. Both $\pi_\theta$ (senior, trainable) and $\pi_{\text{jun}}$ (junior, frozen) are autoregressive: $\pi(y \mid x) = \prod_{t=1}^T \pi(y_t \mid h_t)$. We compare two objectives.

\paragraph{(a) GRPO with KL regularization toward $\pi_{\text{jun}}$.}
\begin{equation}
  \mathcal{L}_{\text{KL-Reg}}(\theta)
  \;=\; -\,\mathbb{E}_{y \sim \pi_\theta}\!\Big[\textstyle\sum_t A_t \log \pi_\theta(y_t \mid h_t)\Big]
  \;+\; \beta \cdot \mathbb{E}_{y \sim \pi_\theta}\!\Big[\textstyle\sum_t \mathrm{KL}\!\big(\pi_\theta(\cdot \mid h_t) \,\big\|\, \pi_{\text{jun}}(\cdot \mid h_t)\big)\Big].
  \label{eq:klreg}
\end{equation}
Rollouts are sampled solo from $\pi_\theta$; the regularizer penalizes per-position next-token disagreement with $\pi_{\text{jun}}$ along senior-natural trajectories. In implementation, we use verl's low-variance KL estimator (\texttt{low\_var\_kl}), with $\beta$ set by \texttt{kl\_loss\_coef}.

\paragraph{(b) \modelname{}.} Let $M_p$ denote the tandem rollout distribution: at each handoff boundary an independent $\mathrm{Bernoulli}(p)$ draw determines the active model $a_t \in \{\text{sen}, \text{jun}\}$, and $y_t \sim \pi_{a_t}(\cdot \mid h_t)$.
\begin{equation}
  \mathcal{L}_{\text{\modelname{}}}(\theta)
  \;=\; -\,\mathbb{E}_{y \sim M_p}\!\Big[\textstyle\sum_{t \,:\, a_t = \text{sen}} A_t \log \pi_\theta(y_t \mid h_t)\Big].
  \label{eq:trl-obj}
\end{equation}

Since only senior-emitted positions depend on $\theta$, the standard policy-gradient identity gives
\begin{equation}
  \nabla_\theta\, \mathbb{E}_{y \sim M_p}[r(y)]
  \;=\; \mathbb{E}_{y \sim M_p}\!\Big[r(y) \textstyle\sum_{t \,:\, a_t = \text{sen}} \nabla_\theta \log \pi_\theta(y_t \mid h_t)\Big].
\end{equation}
Unlike KL-Reg, \modelname{} evaluates the reward on joint rollouts sampled under the tandem mixture schedule. Crucially, $r$ is evaluated on a trajectory that includes junior-emitted tokens, so the team-reward feedback at any senior-emitted position $t$ reflects whether the junior's contributions remain compatible with the rest of the trajectory. KL-Reg, as instantiated here, receives the senior's solo reward and a token-level divergence penalty, but no task reward from trajectories in which the junior participates. This is the additional training signal introduced by tandem rollouts. Consistent with this distinction, among the coefficients evaluated in our ablation, KL-Reg either retains the communication tax of vanilla GRPO or improves junior legibility at the cost of solo and team performance. \modelname{} is the only evaluated arm that preserves GRPO-level solo capability while retaining $99.8\%$ of its own capability under handoff.

\end{document}